\documentclass[shortpaper]{clv3}
\usepackage{hyperref}
\usepackage{xcolor}
\definecolor{darkblue}{rgb}{0, 0, 0.5}
\hypersetup{colorlinks=true,citecolor=darkblue, linkcolor=darkblue, urlcolor=darkblue}
\usepackage{times}  
\usepackage{helvet}  
\usepackage{courier}  
\usepackage{url}  
\usepackage{graphicx}  
\usepackage{array}
\usepackage{multirow}
\usepackage{amsmath}
\usepackage{algorithm,algorithmic}
\usepackage{booktabs}
\newcommand{\cmmnt}[1]{}
\newcommand{\newcite}[1]{\namecite{#1}}
\runningtitle{Scalable Micro-planned Generation of Discourse from Structured Data}
\runningauthor{Anirban Laha}
\title{Scalable Micro-planned Generation of Discourse from Structured Data\historydates{Submission received: 4 January 2019; Revised version received:  10 July 2019; Accepted for publication: 17 September 2019}}
\author{Anirban Laha\thanks{The first three authors have equally contributed to the work}}
\affil{Mila, Universit\'{e} de Montr\'{e}al\\
\texttt{anirbanlaha@gmail.com}}
\author{Parag Jain\textsuperscript{$\ast$}}
\affil{School of Informatics,\\
University of Edinburgh\\
\texttt{parag.jain@ed.ac.uk}}
\author{Abhijit Mishra\textsuperscript{$\ast$}}
\affil{IBM Research\\
\texttt{abhijimi@in.ibm.com}}
\author{Karthik Sankaranarayanan}
\affil{IBM Research\\
\texttt{kartsank@in.ibm.com}}
\begin{document}
\maketitle
\begin{abstract}
We present a framework for generating natural language description from structured data such as tables; the problem comes under the category of \textit{data-to-text} natural language generation (NLG). Modern data-to-text NLG systems typically employ end-to-end statistical and neural architectures that learn from a limited amount of task-specific labeled data, and therefore, exhibit limited scalability, domain-adaptability, and interpretability. Unlike these systems, ours is a modular, pipeline-based approach, and does not require task-specific parallel data. It rather relies on monolingual corpora and basic off-the-shelf NLP tools. This makes our system more scalable and easily adaptable to newer domains.

Our system employs a 3-staged pipeline that: (i) converts entries in the structured data to canonical form, (ii) generates simple sentences for each atomic entry in the canonicalized representation, and (iii) combines the sentences to produce a coherent, fluent and adequate paragraph description through sentence compounding and co-reference replacement modules. Experiments on a benchmark mixed-domain dataset curated for paragraph description from tables reveals the superiority of our system over existing \textit{data-to-text} approaches. We also demonstrate the robustness of our system in accepting other popular datasets covering diverse data types such as \textit{Knowledge Graphs} and \textit{Key-Value} maps.  
\end{abstract}

\section{Introduction}
\label{sec:intro}
Structured data, such as tables, knowledge graphs, or dictionaries containing key-value pairs are popular data representation mechanisms used in a wide variety of industries to capture domain-specific knowledge. As examples, (1) in the \textit{finance} domain, tabular data representing the financial performance of companies, (2) in \textit{healthcare}, information about chemical composition of drugs, patient records \textit{etc.}, (3) in \textit{retail}, inventory records of products and their features, are few among many other manifestations of structured data. Various AI-based human-machine interaction applications such as question-answering or dialog involve retrieving information from such structured data for their end goals. A key component in such applications deals with Natural Language Generation (NLG) from the aforementioned structured data representations, known as the \emph{data-to-text} problem. Another important use-case of this problem is \emph{story-telling} from data, as in automatic report generation. 

\begin{table}[t]
\small
\caption{
Existing Structured data-to-text systems : While \textbf{\textsc{WebNLG} Model} is a system trained on \textsc{WebNLG} training data \cite{nlg-micro17}, \textbf{\textsc{WikiBio} Model} is a system trained on \textsc{WikiBio} training data \cite{lebret2016neural}. As expected, the performance of both these models are good only on the dataset they are trained on, showing lacking of adaptability across domains.}
\label{tab:cross}
\begin{scriptsize}
\begin{tabular}{l p{10.5cm}}
    \toprule
    \multicolumn{2}{c}{\textbf{\textsc{WikiBio} Test data}}\\
    \midrule
    \textbf{Input} & \textbf{Title} : Thomas Tenison \\
        & \textbf{Birth Date:} 29 September 1636 \\
        & \textbf{Birth Place :} Cottenham , Cambridgeshire , England \\
        & \textbf{Death Date :} 14 december 1715 \\
        & \textbf{Death Place :} London , England  \\
        & \textbf{Archbishop of:} Archbishop of Canterbury \\
        & \textbf{Enthroned:} 1695 \\
        & \textbf{Ended :} 14 December 1715 \\
        & \textbf{Predecessor :} John Tillotson \\ 
        & \textbf{Successor :} William Wake\\
    \midrule
    \textbf{Reference:}&Thomas Tenison (29 September 1636 - 14 December 1715) was an English church leader, Archbishop of Canterbury from 1694 until his death.\\
    \textbf{\textsc{WebNLG} Model:}&thomas , england and england are the main ingredients of thomas of archbishop , which is a member of the title of the thomas of archbishop . The birth date of the country is thomas.\\
    \textbf{\textsc{WikiBio} Model:}&thomas tenison (29 september 1636 - 14 december 1715) was archbishop of canterbury from 1695 to 1715. \\
    \midrule
    \multicolumn{2}{c}{\textbf{\textsc{WebNLG} Test data}}\\
    \midrule
    \textbf{Input} & \textbf{Bacon Explosion} , \textit{country} , \textbf{United States}\\
	& \textbf{United States} , \textit{leader name} , \textbf{Barack Obama} \\
	& \textbf{United States} , \textit{ethnic group} , \textbf{White Americans}\\
    \midrule
    \textbf{Reference:}&Bacon Explosion comes from the United States where Barack Obama is the leader and white Americans are an ethnic group.\\
    \textbf{\textsc{WebNLG} Model:}&The Bacon Explosion comes from the United States where Barack Obama is the leader and White Americans are one of the ethnic groups.\\
    \textbf{\textsc{WikiBio} Model:}&bacon explosion is a united states competitive american former competitive men 's national team. united states ( born october 16 , 1951 ) is a retired united states district judge for the united states district court for the united states district court for the united states district court for the united states district court for the united states district court for the united states district court for the united states district court for the united states district court for the united states district court for the united states district court for the united states united states ( born october 16 , 1949 ) is an american former white executive.
\\
    \cmmnt{\midrule
    \multicolumn{2}{c}{\textbf{\textsc{\textcolor{red}{Remove WikiTable}} Test data}}\\
    \midrule
    \textbf{Reference}&\\
    \textbf{\textsc{WebNLG} Model}&\\
    \textbf{\textsc{WikiBio} Model}&\\}
\end{tabular}
\end{scriptsize}
\end{table}

In literature, several approaches have been proposed for \emph{data-to-text}, which can be categorized as \textit{rule based} systems \cite{dale2003coral,reiter2005choosing}, \textit{modular statistical} techniques \cite{barlapata05,konstas2013inducing} and more recently, \textit{end-to-end neural} architectures \cite{lebret2016neural,weather16,N18-2098,N18-1139}. Rule-based approaches employ heuristics or templates for specific tasks which cannot scale to accommodate newer domains unless heuristics are revised manually. 
On the other hand, the statistical and neural approaches require large amounts of parallel labeled data for training. Parallel data in NLG tasks are quite expensive to obtain; they require an annotator to frame a complete text as output for each input. To work on unseen domains and tasks, these data-hungry systems need to be trained again with parallel data for every new domain. To put this in the \textit{data-to-text} NLG perspective, \textit{Table} \ref{tab:cross} shows lack of adaptability of supervised systems on unseen domain data. It can be seen that models do well for the domain in which they are trained on while they perform poorly on a different domain. In hindsight, such end-to-end systems are adversely affected by even slight changes in input vocabulary and may not generate language patterns other than what is seen during training.

Further, since existing systems are designed as task-specific solutions, they tend to jointly learn both \emph{content selection} from the input (what to say?) and the \emph{surface realization} or language generation (how to say?). This is often undesirable as the former, which decides ``what is interesting'' in the input, can be highly domain-specific. For example, what weather parameters (temperature, wind-chill) are influential versus what body parameters (heart-rate,  body temperature) are anomalous are heavily dependent on the domain at hand, such as weather or healthcare respectively. Whereas the latter part of language generation may not be as much domain dependent and can, thus, be designed in a reusable and scalable way. Therefore, it would be easier to develop scalable systems for language generation independently than developing systems that jointly learn to perform both content selection and generation.

In this article, we propose a general purpose, unsupervised approach to language generation from structured data; our approach works at the linguistic level using word and sub-word level structures.
The system is primarily designed for taking a structured table with variable schema as input and producing a coherent paragraph description pertaining to the facts in the table. However, it can also work with other structured data formats such as graphs and key-value pairs (in the form of JSONs) as input. Multiple experiments show the efficacy of our approach on different datasets having varying input formats without being trained on any of these datasets. By design, the system is unsupervised and scalable, \textit{i.e.} it assumes no labeled corpus and only considers monolingual, unlabeled corpora and WordNet during development, which are inexpensive and relatively easy to obtain. 

In the proposed approach, the generation of description from structured data happens in three stages, \textit{viz.} (1) \textbf{canonicalization}, where the input is converted to a standard canonical representation in the form of tuples, (2) \textbf{simple language generation}, where each canonical form extracted from the input is converted into a simple sentence, and (3) \textbf{discourse synthesis and language enrichment}, where simple sentences are merged together to produce complex and more natural sentences. The first stage is essential to handle variable schema and different formats. The second stage gleans morphological, lexical, and syntactic constituents from the canonical tuples, and stitches them into simple sentences. The third stage applies sentence compounding and co-reference replacement on the previously produced simple sentences to generate a fluent and adequate description. For the development of these modules, at most a monolingual corpus, WordNet, and three basic off-the-shelf NLP tools namely, part-of-speech tagger, dependency parser and named entity recognizer are needed.

To test our system, we first curate a multi-domain benchmark dataset (referred henceforth as \textbf{\textsc{WikiTablePara}}) that contains tables and corresponding manually written paragraph descriptions; and to the best of our knowledge, such a dataset does not exist. Our experimental results on this dataset demonstrate the superiority of our system over the existing \emph{data-to-text} systems. Our framework can also be extended to different schema and datatypes. To prove this, we perform additional experiments on two datasets representing various domains and input-types, only using their test splits: (i) \textsc{WikiBio}~\cite{lebret2016neural}, representing key-value pairs, and (ii) \textsc{WebNLG}~\cite{nlg-micro17}, representing knowledge graphs. Additionally, for the sake of completeness, we extend our experiments and test our system's performance on existing data-to-text NLG datasets (for the task of tuple to text generation). We demonstrate that even though our system does not undergo training on any of these datasets, it nevertheless delivers promising performance on their test splits. The key contributions of this article are as summarized below:
\begin{itemize}
    \item We propose a general purpose, unsupervised, scalable system for generation of descriptions from structured tables with variable schema and diverse formats.
    \item Our system employs a modular approach enabling interpretability, as the output of each stage in our pipeline is in a human-understandable textual form. 
    \item We release a dataset called \textbf{\textsc{WikiTablePara}} containing WikiTables and their descriptions for further research. Additionally, we also release data gathered for  modules for sentence realization from tuples (refer \textit{Secs.} \ref{sec:simplegen}, and \ref{sec:datasets}, useful for general purpose tuple/set to sequence tasks. The dataset and code for our experiments are available at \url{https://github.com/parajain/structscribe}.
\vspace{-0.35cm}
\end{itemize}

We would like to remind our readers that our system is unsupervised as it does not require parallel corpora containing structured data such as tables at the source side end and natural language description at the target side. Manually constructing such labeled data can be more demanding than some of the well-known language generation tasks (such as summarization and translation) because of the variability of the source structure and the non-natural association between the source and target sides. Our system does not require such parallel data and divides the problem into sub-problems. It requires only simpler data forms that can be curated from unlabeled sources. 

We would also like to point that an ideal description generation system would require understanding the pragmatic aspects of the structure under consideration. Incorporating pragmatic knowledge still remains an open problem in the domain of NLG, and our system's capability towards handling pragmatics is rather  limited. As the state-of-the-art progresses, we believe that a modular approach such as the one proposed can be upgraded appropriately.

\section{Central Challenges and Our Solution}
\label{sec:challenges}
\begin{figure}[t]
\includegraphics[width=13cm, height=8cm, keepaspectratio=true]{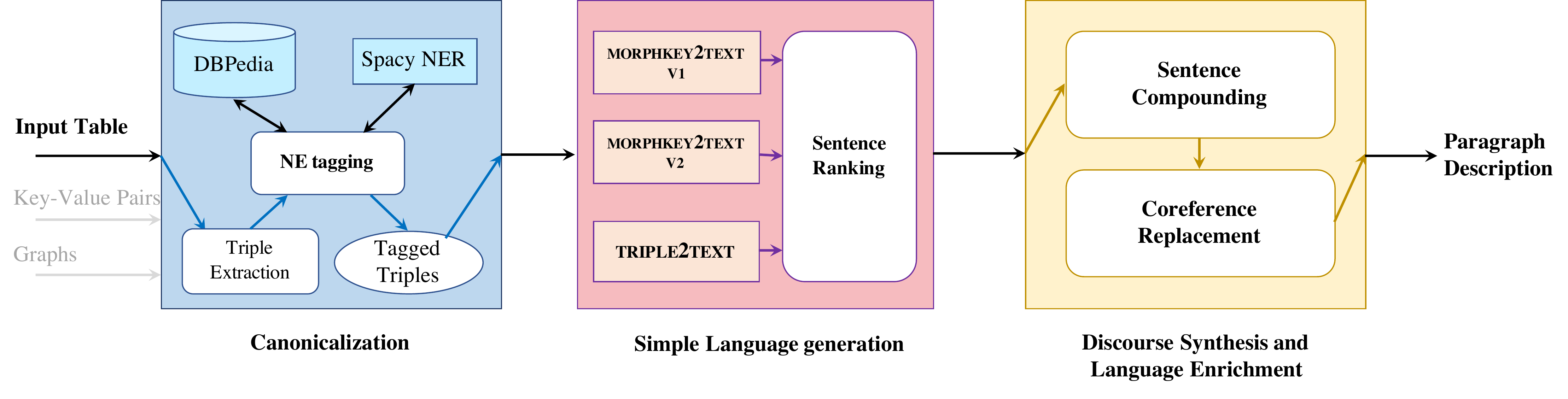}
\caption{Our proposed 3-staged modular architecture for description generation from structured data}
\label{fig:system}
\end{figure}

This section summarizes the key challenges in description generation from structured data.
\begin{itemize}
    \item \textbf{Variable Schema:} Tables can have variable number of rows and columns. Moreover, the central theme around which the description should revolve can vary. For example, two tables can contain  column-headers \emph{[Company Name, Location]}, yet the topic of the description can be the \textit{companies} or the \textit{locations} of various companies. Also, two tables having column-headers \emph{[PlayersName, Rank]} and \emph{[Rank, PlayersName]} represent the same data but may be handled differently by existing-methods that rely on ordered-sequential inputs. 
    \item \textbf{Variation in Presentation of Information:} The headers of tables typically capture information that is crucial for generation. 
    However, presentations of headers can considerably vary for similar tables. For example, two similar tables can have column-headers like \emph{[Player, Country]} and \emph{[Player Name, Played for Country]}, where the headers in the first table are single-word nouns but the first header of the second table is a \textit{noun-phrase} and the second header is \textit{verb-phrase}. It is also possible that the headers share different inter-relationships. Nouns such as \emph{[Company, CEO]} should represent the fact that CEO is a part of the company, whereas entities in headers \emph{[temperature, humidity]} are independent of each other. 
    \item \textbf{Domain Influence:} It is known that changing the domain of the input has adverse effects on end-to-end generators, primarily due to differences in vocabulary (\textit{e.g.,} the word ``tranquilizer'' in healthcare data may not be found in tourism data).
    \item \textbf{Natural Discourse Generation:} Table descriptions in the form of discourse (paragraphs) should contain a natural flow with a  mixture of simple, compound, complex sentences. Repetition of entities should also be replaced by appropriate co-referents. In short, the paragraphs should be fluent, adequate and coherent.
\end{itemize}
End-to-end neural systems mentioned in the previous sections suffer from all the above challenges. According to \newcite{nlg-micro17}, these systems tend to overfit the data they are trained on, \textit{``generating domain specific, often strongly stereotyped text''} (\textit{eg.} weather forecast or  game commentator reports). Rather than learning the semantic relations between data and text, these systems are heavily influenced by the style of the text, the domain vocabulary, input format of the data and co-occurrence patterns. As per \newcite{D17-1239}, \textit{``Even with recent ideas of copying and reconstruction, there is a significant gap between neural models and template-based systems, highlighting the challenges in data-to-text generation''}. Our system is designed to address the challenges to some extent through a three-staged pipeline, namely, (a) \textbf{canonicalization}, (b) \textbf{simple language generation}, and (c) \textbf{discourse synthesis and language enrichment}. In the first stage, the input is converted to a standard canonical representation in the form of tuples. In the second stage, each canonical form extracted from the input is converted to simple sentences. In the final stage, the simple sentences are combined to produce coherent descriptions. The overall architecture is presented in \textit{Fig.} \ref{fig:system}. 

Note that our pipeline is designed to work with tables which do not have a hierarchy amongst its column headers and row headers. We believe that tables of such kind can be normalized as a pre-processing step and then fed to our system. To handle this pre-processing is beyond the scope of the current work. We discuss our central idea in the following sections.
\section{Canonicalization of Structured Data}
\label{sec:canonical}
Our goal is to generate descriptions from structured data which can appear in various formats. For this, it is essential to convert the data to a canonical form which can be handled by our generation stages. Though our main focus is to process data in tabular form, the converter is designed to handle other input formats as well, as discussed below.
\subsection{Input Formats}
\begin{enumerate}
    \item \emph{Table}: Tables are data organized in rows and columns. We consider single-level row and column headers with no hierarchy. A table row can be interpreted as an $n$-ary relation. Currently, we simplify table row representation as a collection of binary relations (or triples).
    \item \emph{Graph}: Knowledge Graphs have entities represented as nodes while edges denote relations between entities. Here we consider only binary relations. A knowledge graph can be translated as a collection of binary relations or triples.
    \item \emph{JSON}: This is data organized in the form of a dictionary of key-value pairs. We limit ourselves to single-level key-value pairs where the keys and values are literals. A pair of key-value pairs are converted to a triple by concatenating the value term of the first key-value pair with the second key-value pair.
\end{enumerate}
\subsection{Canonical Form and Canonicalization}
\label{sec:canon}
For our system to be able to handle various formats listed above, we need to convert them to a standard format easily recognizable by our system. Moreover, it is required that the generation step can be trained without involving labeled parallel data so that they can be used in various domains where only monolingual corpora is available. Keeping this in mind, we arrived at a canonical form consisting of triples made of binary relations among two entities types. For example, consider the triple : $\langle$\emph{Albert Einstein ; birth place ; Ulm, Germany}$\rangle$. The entity tags for named entities `Albert Einstein' and `Ulm, Germany' are PERSON and GPE respectively. This leads to a canonicalized triple form as the following
\begin{align*}
    \langle\textbf{PERSON} \ \text{birth place} \ \textbf{GPE}\rangle 
\end{align*}

For tabular inputs, extraction of tuples require the following assumption to be followed. 
\begin{itemize}
 \item The column-headers of the table should be considered as the list of keywords that decide the structure of the sentences to be generated. In case the table is centered around row headers (\textit{i.e.,} row headers contain maximum generic information about the table), the table has to be transposed first.
 \item One column header is considered as the \textit{primary key}, around which the theme of the generated output revolves. For simplicity, we chose the first column-header of the tables in our dataset to be the primary key.
\end{itemize}

\begin{figure}[t]
\centering
\includegraphics[width=0.9\textwidth, height=7cm, keepaspectratio=true]{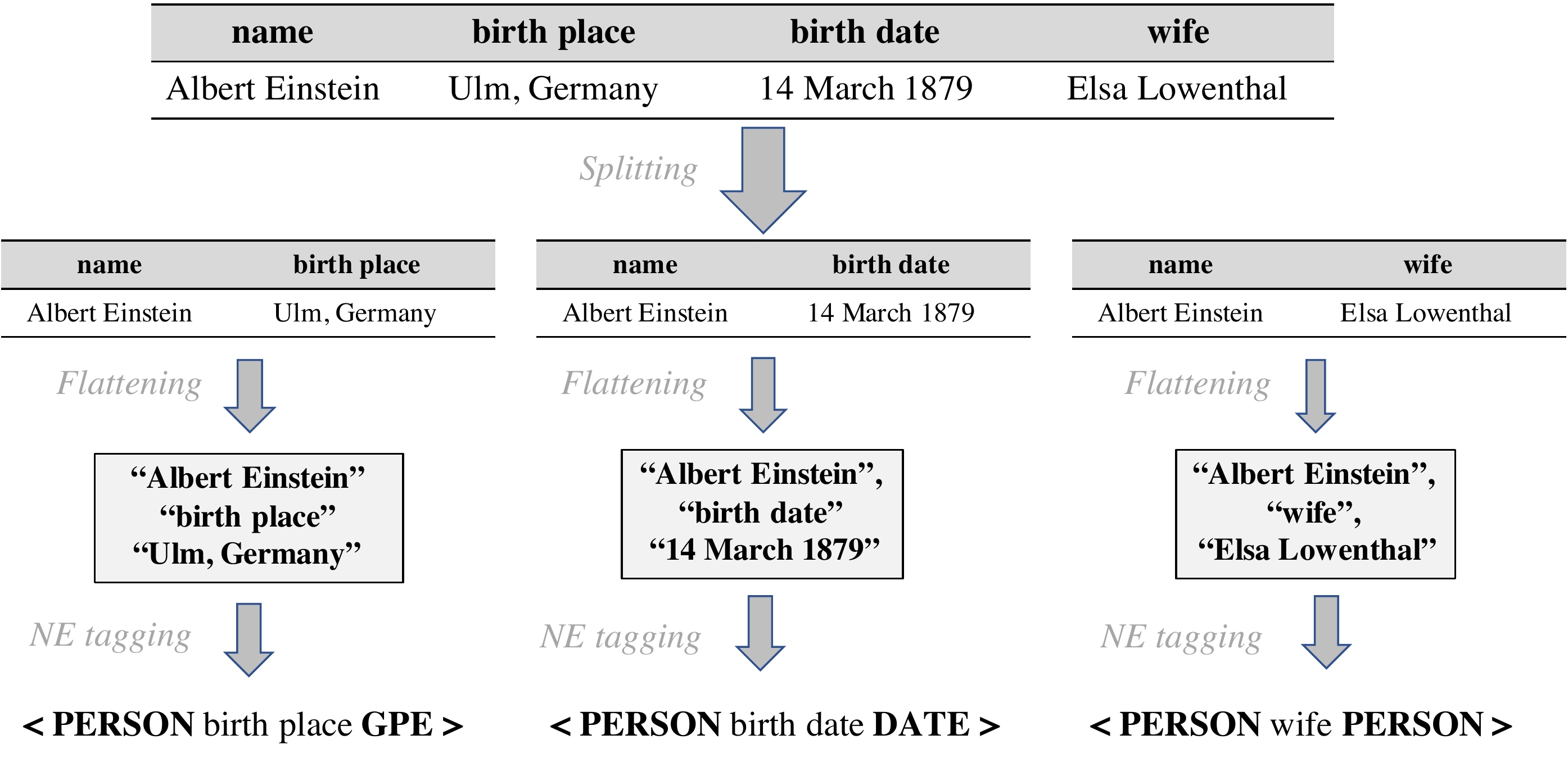}
\caption{Example of extraction of canonical triples from tabular inputs}
\label{fig:canonical}
\end{figure}

For each table, the table is first broken into a set of subtables containing 1-row and 2-columns, as shown in \textit{Fig.} \ref{fig:canonical}. The first columns of the subtables represent the primary-key of the table. For a table containing $M$ rows and $N$ columns (excluding headers), a total number of $M \times (N-1)$ subtables are thus produced. The subtables are then flattened to produce a triple by dropping the primary key header and concatenating the entries of the subtables, as shown in \textit{Fig.} \ref{fig:canonical}. This produces standard entity-relationship triples $\langle e_1,r,e_2\rangle$ where $e_1$, and $e_2$ are entities that are entries and $r$ is the relationship, which is captured by the column header.

The entities $e_1$, and $e_2$ are tagged using an NER-tagger, which assigns domain-independent place-holder tags such as \textbf{PERSON} and \textbf{GPE} for persons and geographical regions respectively. For tagging we use Spacy (\texttt{spacy.io}) NER tagger, an off-the-shelf tagger that performs reasonably well even on words and phrases. We also employ a DBPedia lookup\footnote{Refer \url{https://github.com/dbpedia/lookup}} based on exact string matching in situations where NER is unable to recognize the named entity. String matching is done with either the URI-labels or the  anchor-texts referring to the URI to find out the relevant tag. This is helpful in detection of peculiar multi-word named entities like `The Silence of the Lambs' which will not be recognized by Spacy due to lack of context. All DBPedia classes have been manually mapped to 18 Spacy NER types. As a fallback mechanism, any entity not recognized through DBPedia lookup is assigned with \textbf{UNK} tag. This process produces somewhat domain independent canonical representations from the tables as seen in \textit{Fig.} \ref{fig:canonical}. The NER tags and the corresponding original entries are carried forward and remain available for use in stages $2$ and $3$. At stage $2$, these tags are replaced with the original entries to form proper sentences. The tags and the original entries are also used for language enrichment in stage $3$.

Unlike tables, for input types like knowledge graphs and key-value pairs, extraction of canonical triples is straightforward. Knowledge graphs  typically follow the triple form with nodes representing entities and edges representing relations. Similarly, a pair of key-value entries can be flattened and a triple can be extracted. All these formats, thus, can be standardized to a collection of canonical triples with NE tags acting as placeholders. 

In the following section, we describe how a simple sentence can be extracted from each canonical triple. A collection of canonical triples obtained from a table (or other input types) will produce a collection of simple sentences, which is compounded to form a coherent description.
\section{Simple Language Generation}
\label{sec:simplegen}
The simple language generation module takes each canonical triple and generates a simple sentence in natural language. For instance, the triple $\langle$\textbf{PERSON} \ \text{birth place} \ \textbf{GPE}$\rangle$ will be translated to a simple sentential form like the following: \cmmnt{ $\langle$\textbf{PERSON} \emph{was born in} \textbf{GPE}$\rangle$.}
\begin{align*}
    \langle\textbf{PERSON} \ \emph{was born in} \ \textbf{GPE}\rangle 
\end{align*}
This will finally be replaced with the original entities to produce a simple sentence as follows: \emph{Albert Einstein was born in Ulm,  Germany}. The canonical triple set in \emph{Fig.} \ref{fig:canonical} should produce the following (or similar) simple sentences (refer as set \ref{eqn:simple}):
\begin{align}
    \textit{Albert Einstein was born in Ulm,  Germany} \nonumber\\
    \textit{Albert Einstein has birthday on 14 March 1879} \nonumber\\
    \textit{Elsa Lowenthal is the wife of Albert Einstein}
\label{eqn:simple}
\end{align}

This is achieved by the following steps : \textbf{(1) Preprocessing} - which transforms the canonical triple to a modified canonical triple, \textbf{(2) TextGen} - which converts the modified canonical triple to a simple sentential form like $\langle$PERSON \emph{was born in} GPE$\rangle$, \textbf{(3) Postprocessing} - which replaces back the original entities to produce a simple sentence like \emph{Albert Einstein was born in Ulm,  Germany}, and lastly, \textbf{(4) Ranking} - which selects the best sentence produced in step 3 when multiple variants of TextGen are run in parallel. The details of  these steps are shared below.

\subsection{Preprocessing}
It is possible that the canonical triples will contain words that cannot be easily converted to a sentence form without additional explicit knowledge. For example, it may not be easy to transform the vanilla triple $\langle$PERSON \ \text{game} \ Badminton$\rangle$ to a syntactically correct sentence  $\langle$PERSON \ \textit{plays} Badminton$\rangle$.

To convert the relation term into a verb phrase we employ a pre-processing step. The step requires two resources to be available - (1) WordNet and (2) Generic Word embeddings, at least covering the default vocabulary of the language (English). We use the 300-dimensional \texttt{glove} embeddings for this purpose \cite{pennington2014glove}.

The preprocessing step covers the following two scenarios:
\begin{enumerate}
    \item \textbf{Relation term is a single-word term:} In this case, the word is lemmatized and the root form is looked up in a verb lexicon pre-extracted from WordNet. If the look up succeeds, the lemma form is retained in the modified triple. Otherwise, the top $N$ verbs\footnote{$N$ is set to $10$ in our setup.} that are closest to the word are extracted using \texttt{glove} vector based \textit{cosine similarity}. For example, through this technique, for the original word ``game'', which is not a verb, related verbs such as ``match'' and ``play'' can be extracted. The verb ``play'' will be the most suitable one for generating a sentence later. The most suitable verb is decided as follows. For each extracted verb $v$ related to the original word $o$, the $synsets$ for $v$ and $o$ are extracted from WordNet. The glosses and examples for each synset of $o$ are extracted from WordNet and combined to form a textual representation ($F_o$). Similarly, the textual representation ($F_v$) considering the glosses and examples of synsets of $v$ is formed. The \textit{degree of co-occurrence} of words $v$ and $o$ is computed using the normalized  counts of co-occurrences of $v$ and $o$ in $F_o$ and $F_v$. The candidate verb having the highest \textit{degree of co-occurrence} is selected as the most appropriate verb. Through this, the word ``play'' would be selected as the most  appropriate verb for the word ``game'', as both words will co-occur in the glosses and examples of synsets of both ``game'' and ``play''. 
    \item \textbf{Relation term is a multi-word term:} The relation term, in this case, would contain both content (\emph{i.e.} non-stopwords) and function words (\emph{i.e.} stopwords). Examples of multi-word terms are ``\textit{country played for}'' and ``\textit{number of reviews}''. When such terms are encountered, the main verb in the phrase is extracted through \textit{part-of-speech} (POS) tagging. If a verb is present, the phrase is altered by moving the noun phrase preceding the verb to the end of the phrase. So, the phrase ``country played for'', through this heuristic, would be transformed to ``played for country''. This is based on the assumption that in tabular forms, noun phrases that convey an \textit{action} are actually a transformed version of a verb phrase. 
\end{enumerate}

The above preprocessing techniques modify the input triple  which we refer to as \emph{modified canonical triple}. This step is useful for the \textsc{triple2text} generation step as discussed next.
\vspace{-0.2cm}
\subsection{TextGen}
\label{sec:triple2text}
The objective of this step is to generate simple and syntactically correct sentences from the (modified) canonical triples. 
We propose three ways to generate sentential forms as elaborated below. All the below mentioned ways are different alternatives to generate a simple sentential form, hence they can be executed in parallel.
\subsubsection{\textsc{triple2text}}
This module is the simplest and is developed using a \texttt{seq2seq} \cite{opennmt} network, which is trained on the curated \textsc{Triple2Text} dataset (refer \textit{Sec.} \ref{sec:datasets} Dataset 3). The dataset consists of triples curated from various sources of knowledge bases extracted from open web-scale text dumps using popular information extraction techniques (such as \newcite{banko2007open,schmitz2012open}). Additionally, existing resources such as Yago Ontology\cite{suchanek2007yago} and VerbNet\cite{Schuler:2005:VBC:1104493} are employed. The criteria for constructing triples and simple sentence pairs (used as target for training \texttt{seq2seq}) are different for different resources. We should point to our readers that no annotation was needed for creation of this dataset as the simple sentences were constructed by concatenation of elements in the triples (discussed in \textit{Sec.} \ref{sec:datasets} Dataset 3). Only this variant of generation requires a modified canonical triple, obtained using the preprocessing step mentioned above. The other variants can work with the canonical triple without such modification.
\begin{figure}[t]
\centering
\includegraphics[width=12cm, height=10cm, keepaspectratio=true]{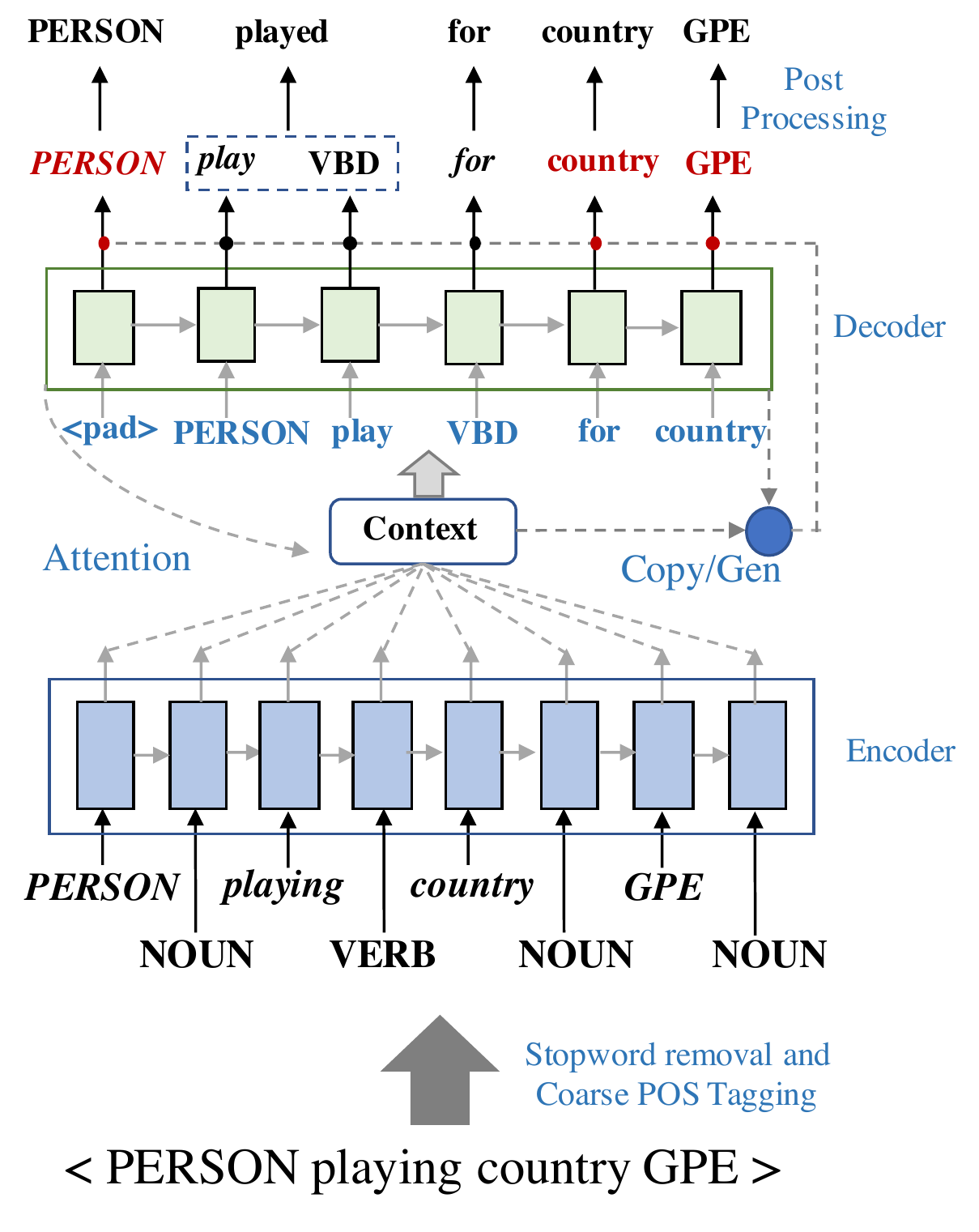}
\caption{Example demonstrating the working of the \textsc{morphkey2text} TextGen system. Generated words shown in red color are produced via copy operation and those in black color are produced via generation operation.}
\label{fig:morph}
\end{figure}
\subsubsection{\textsc{morphkey2text} (\textsc{v1} and \textsc{v2})} 
\label{sec:morphkey2text}
The conversion of any canonical triple to a sentence demands the following linguistic operations: 
\begin{enumerate}
    \item determining the appropriate morphological form for the words/phrase in the canonical triple, especially the relation word/phrase (\textit{e.g.,} transforming the word ``play'' to ``played'' or ``plays'').
    \item determining the articles and prepositions necessary to construct the sentences (\textit{e.g.,} transforming ``play'' to ``plays for''). 
    \item adding appropriate auxiliary verbs when necessary. This is needed especially for passive forms (\textit{e.g.,} transforming ``location'' to ``is located at'' by adding the auxiliary verb ``is'').
\end{enumerate}
Ideally, any module designed for canonical-triple to sentence translation should dynamically select a subset of the above operations based on the contextual clues present in the input. To this, we propose the  \textsc{morphkey2text} module, a variant of \texttt{seq2seq} network empowered with attention and copy mechanisms. \textit{Fig.} \ref{fig:morph} shows a working example of the \textsc{morphkey2text} system. We skip explaining the well-known seq2seq framework for brevity. As input, the module takes a processed version of the canonical triple in which (a) NE tags are retained (b) Stopwords are removed if they appear in the relation terms in the canonical triples and (c) The coarse POS tags for both the NE tags and words are appended to the input sequence. The module is expected to generate a sequence of words along with the fine-grained POS tags (in PENN tagset format) for the verbs appearing in their \textit{lemma} form. The rationale behind such an input-output design is that, \textbf{dealing with the lemma forms at the target side and incorporating additional linguistic signals in terms of POS should enable the system to apply appropriate changes at morphological and lexical levels}. This will, in turn, help address the problem of lexical and morphological data-sparsity across domains better. As seen in \textit{Fig.} \ref{fig:morph}, the canonical triple $\langle$\textbf{PERSON}  \emph{playing country} \textbf{GPE}$\rangle$, is first transformed into a list of content words and their corresponding coarse-grained POS tags. During generation, the input key-word and POS \emph{``playing VERB''} are translated to \emph{``play VBD''} and the output is post-processed to produce the word \emph{``played''}. As the system has to deal with lemma forms and NE and POS tags at both input and output sides, it allows the system to just \emph{copy} input words, which makes the system robust across domains.

\begin{figure}[ht]
\centering
\includegraphics[width=12cm, keepaspectratio=true]{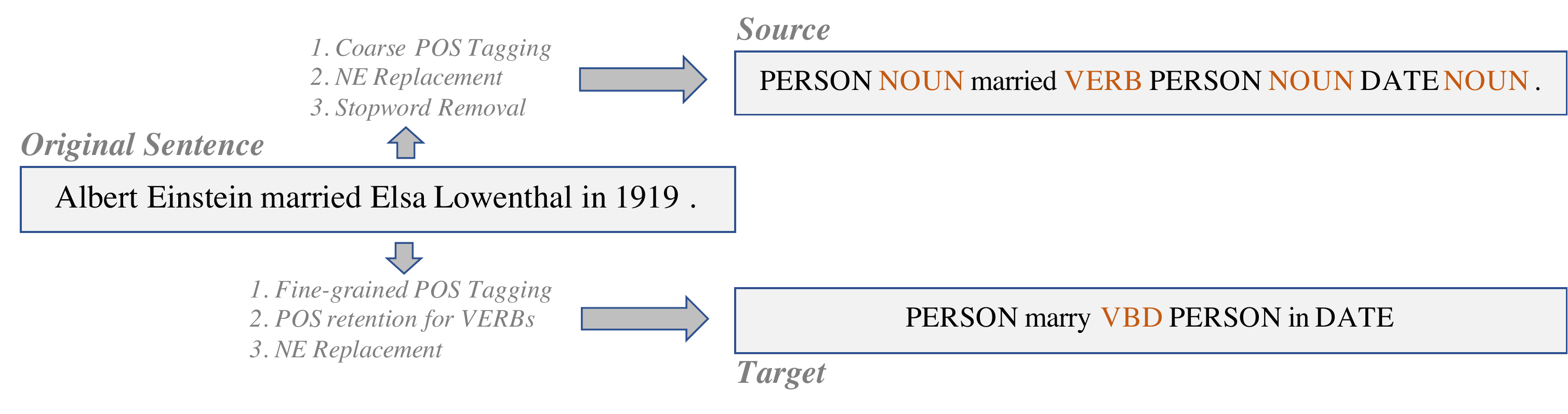}
\caption{Extraction of a single training instance from an unlabeled sentence for the  \textsc{morphkey2text} TextGen system.}
\label{fig:morphexample}
\end{figure}

Preparing training data for the \textsc{morphkey2text} design requires only a monolingual corpus and a few general purpose NLP tools and resources such as  POS tagger, NE Tagger, and WordNet. A large number of simple sentences extracted from web-scale text dumps (such as Wikipedia) are first collected. The sentences are then POS tagged and the named entities are replaced with NE tags. Stopwords (function words) such as articles and prepositions are dropped from the sentences by looking up in a stopword lexicon. Since the POS tagger produces fine-grained POS-tags, the tags are converted to coarse POS tags using a predefined mapping.  This produces the \textit{source} (input) side of the training example. As of \textit{target} (output), the named entities in the original sentences are replaced with NE tags, the other words are \textit{lemmatized} using WordNet lemmatizer, and the fine-grained POS tags of the words are augmented if the lemma form is not the same as the base form. \textit{Fig.} \ref{fig:morphexample} illustrates construction of a training example from unlabeled data.

We implement two different variants of the \textsc{morphkey2text} system. The \textsc{morphkey2text V1} module is trained based on the \textsc{MorphKey2Text} dataset (version v1) that was created from monolingual corpora (explained in  \emph{Sec.} \ref{sec:datasets} Dataset 2). The \textsc{morphkey2text v2} is trained on a different version (v2) of the \textsc{MorphKey2Text} dataset (details in \emph{Sec.} \ref{sec:datasets} Dataset 2).
\subsection{Post-processing} This step restores the original entities from the input by replacing the tagged forms generated from the step above. Additionally, if possessive nouns are detected in the sentence, apostrophes are added to such nouns. Possessives are checked using the following heuristic -  \textit{if the POS tag for the word following the first entity is not a verb, the word is a potential possessive candidate}. Postprocessing is applied to each of the competing modules enlisted in the above step.

The above variants \textsc{triple2text}, \textsc{morphkey2text v1} and \textsc{morphkey2text v2} can run in parallel to produce different translations of the canonical triple. Out of these, the best produced sentence are selected by the ranking step mentioned below.
\subsection{Scoring and Ranking}
To select the most appropriate output from the TextGen systems discussed earlier, a \textbf{ranker} is employed; it sorts the sentence based on a composite score as given below:
\begin{equation}
score (i,s) = f(s) \times g(i,s)
\label{eq:score}
\end{equation}

where $i$ and $s$ represent the canonical triple and generated sentence. Functions $f(.)$ and $g(.)$ represent the \textit{fluency} (grammaticality) of the output sentence and \textit{adequacy} (factual overlap between input and output). The \textit{fluency} function $f$ is defined as follows:
\begin{equation}
f(s) = LM (s)  = LM (w_1,w_2,...,w_N)
\label{eq:fluency}
\end{equation}

where for a sentence of N words ${w_1,w_2,...,w_N}$, the LM, an N-gram language model, returns the likelihood of the sentence. For this, a 5-gram general purpose language model is built using Wikipedia dump and KenLM \cite{heafield2011kenlm}. The \textit{adequacy} function $g$ is defined as:
\begin{equation}
g(i,s) = \frac{\# co\text{-}occurring~words~in~i~and~s}{\# words~in~i} 
\label{eq:adequacy}
\end{equation}

Before applying the ranker, we employ heuristics to filter incomplete and un-natural sentences. Sentences without verbs or entities and sentences that are disproportionately larger or smaller than the input are discarded. Once the ranker produces the best simple sentence per input triple, the simple sentences are then combined into a coherent paragraph as explained below.
\section{Discourse Synthesis and Language Enrichment}
\label{sec:discourse}
In this section, we discuss how to combine the collection of generated simple sentences set \ref{eqn:simple} from \emph{Sec.} \ref{sec:simplegen} to produce a paragraph by merging sentences as shown below:

\begin{quote}
    \textit{Albert Einstein was born in Ulm,  Germany and has birthday on 14 March 1879. Elsa Lowenthal is the wife of Albert Einstein.}
\end{quote}

The above paragraph is produced by a \emph{sentence compounding module} followed by a \textit{coreference replacement module} to produce the final coherent paragraph: 
\begin{quote}
    \textit{Albert Einstein was born in Ulm, Germany and has birthday on 14 March 1879. Elsa Lowenthal is the wife of him}.
\end{quote}
\subsection{Sentence Compounding}
This module takes a pair of simple sentences and produces a compound or complex sentence. Every simple sentence is split into a $\langle e_1, rvp, e_2\rangle$ form where $e_1$ and $e_2$ are entities that appear in the input and $rvp$ is the relation verb phrase.
\algsetup{indent=2em}
\newcommand{\compounding}{\ensuremath{\mbox{\textsc{Compound}~($s_1$,$s_2$,$D$)}}}
\begin{algorithm}[t]
\footnotesize
\caption{$\compounding$}
\label{alg:compounding}
\begin{algorithmic}[1]
\STATE $e_{11}$,${rvp}_1$,$e_{12}$ $\gets$ SplitIntoTuple (Sentence $s_1$, $D$)
\STATE $e_{21}$,${rvp}_2$,$e_{22}$ $\gets$ SplitIntoTuple (Sentence $s_2$, $D$)
\STATE {\bf REM} The function \textit{SplitIntoTuple} splits a sentence $s$ into a triple by considering the entities and their corresponding types, the mapping is provided by dictionary $D$.
\IF{$e_{11}~=~e_{21} \And {rvp}_1~=~{rvp}_2$}
    \STATE {\bf REM} \textit{e.g.,} \textit{Jordan played basketball and football.}
    \RETURN  ``$e_{11}~{rvp}_1~e_{12}~\text{and}~e_{22}$''
\ELSIF {$e_{11}~=~e_{21}$}
    \STATE {\bf REM} \textit{e.g.,} \textit{Jordan played basketball and represented U.S.A.}
    \RETURN  ``$e_{11}~{rvp}_1~e_{12}~\text{and}~{rvp}_2~e_{22}$''
\ELSIF{$e_{12} = e_{22} \And {rvp}_1 = {rvp}_2$}
    \STATE {\bf REM} \textit{e.g.,} \textit{Jordan and Kurt played basketball.}
    \RETURN  ``$e_{11}~\text{and}~e_{21}~{rvp}_1~e_{22}$''
\ELSIF {$e_{12} = e_{22}$}
    \STATE {\bf REM} \textit{e.g.,} \textit{Jordan loved and Kurt hated basketball.}
    \RETURN  ``$e_{11}~{rvp}_1~\text{and}~e_{21}~{rvp}_2~e_{22}$''
\ELSIF{$e_{12} = e_{21} \And \text{TypeOf}(e_{12}) = PERSON$}
    \STATE {\bf REM} \textit{e.g.,} \textit{Jordan married Prieto who is a model from Cuba.}
    \RETURN  ``$e_{11}~{rvp}_1~e_{12}~\text{who}~{rvp}_2~e_{22}$''
\ELSIF{$e_{12} = e_{21}$}
    \STATE {\bf REM} \textit{e.g.,} \textit{Jordan played basketball which featured in movie Space Jam.}
    \RETURN  ``$e_{11}~{rvp}_1~e_{12}~\text{which}~{rvp}_2~e_{22}$''
\ENDIF
\end{algorithmic}
\end{algorithm}
For a pair of sentences, if both sentences share the same first entity $e_1$ or both have the same second entity $e_2$, the compounded version can be obtained by `AND'-ing of the relation phrases. In cases where the second entity of one matches the first entity of the following sentence, then a clausal pattern can be created by adding \textit{``who''} or \textit{``which''}. In all other cases, the sentences can be merged by `AND'-ing both the sentences. \emph{Alg.} \ref{alg:compounding} elaborates on this heuristic. This module can also generate different variations of paragraphs based on different combinations of sentences.

\subsection{Coreference Replacement}
\label{sec:coref}
To enhance paragraph coherence, it is often desirable to replace entities that repeat within or across consecutive sentences with appropriate coreferents. For this, we employ a heuristic that  replaces repeating entities with pronominal anaphora.

If an entity is encountered twice in a sentence or appears in consecutive sentences, it is marked as a potential candidate for replacement. The number and gender of the entity are decided using POS tags and an off-the-shelf Gender Predictor module. The module is a CNN-based classifier that trains on person names gathered from various websites. The entity's role is determined based on whether it appears to the left of the verb (\textit{i.e.,} \textit{Agent})) or to the right (\textit{\textit{Object}}). 
\cmmnt{Additionally, \textcolor{blue}{whether the entities with possessives are marked separately.}} Based on the gender, number, role and possessives, the pronouns (he/she/their/him/his \textit{etc.}) are selected, and they replace the entity. We ensure that we replace only one entity in a sentence to avoid incoherent construction due to multiple replacements in close proximity. 

\cmmnt{The three different stages were individually discussed in detail in this section. Please refer \emph{Fig.} \ref{fig:system} for an illustration of the whole proposed system from end to end.} 

We remind our reader about \textit{Fig.} \ref{fig:system} that presents an overview of our system. Due to its modular nature, our system enjoys interpretability; each stage in the pipeline is conditioned on the output of the previous stage. Moreover, all the modules, in principle, can adapt to newer domains. The datasets used for training do not have any domain-specific characteristics and thus these modules can work well across various domains as will be seen in the ``experiments'' section. The whole pipeline can be developed without any parallel corpora of structured table to text. Any data used for training any individual module can be curated from monolingual corpora. The subsequent section discusses such datasets in detail.
\section{Datasets}
\label{sec:datasets}
The section discusses three datasets; Dataset 1 contains tables from various domains and their summaries and can be used for benchmarking any table descriptor generator. Dataset 2 and 3 are developed to train our TextGen modules (\textit{Sec.} \ref{sec:simplegen}). These datasets can be downloaded for academic use from \url{https://github.com/parajain/structscribe}. We also release the code and resources to create similar datasets in a larger scale.

\subsection{Dataset 1: Descriptions from WikiTable (\textsc{WikiTablePara})} We prepare a benchmark dataset for multi-sentence description generation from tables. For gathering input tables, we rely on the \textsc{WikiTable} dataset \cite{P15-1142}, which is a repository of more than 2000 tables. Most of the tables still suffer from the following issues: (a) they do not provide enough context information, as they were originally a part of a Wikipedia page, (b) they are concatenations of multiple tables, and (c) they contain noisy entries. After filtering such tables, we extract 171 tables. Four reference descriptions in the form of paragraphs were manually generated. The average number of sentences for each description in each reference is 12 and the average number of words is between 740-780 respectively. The descriptions revolve around one column of the table, which acts as the primary-key. 
   
\subsection{Dataset 2: Morphological Variation based Keywords-to-Text (\textsc{MorphKey2Text})} 
This is created from monolingual corpora released by \cite{Thorne18Fever}, which is a processed version of Wikipedia dump. We create the first version of the dataset following the technique discussed in \textit{Sec.} \ref{sec:morphkey2text}, \textit{paragraph}. 3, using POS- and NE  taggers.

The second version \textsc{v2} is slightly different in the sense that it employs a higher-recall oriented entity tagging mechanism with the help of POS tags and dependency parse trees of sentences. This is necessary as there are entities such as \textit{``A Song of Ice and Fire''}, which will not be recognized by the NE tagger used to create \textsc{v1}. Such multi-word entities can be detected by a simple heuristic which looks for a sequence of proper nouns (in this case `Song', `Ice' and `Fire') surrounded by stop-words but do not include any punctuation. Moreover, it should not have any verb marked as \texttt{root} by the dependency parser. Through this technique, it is also possible to handle cases where an entity such as \textit{``Tony Blair''} gets detected as two entity tags PERSON and UNK by popular NE taggers such as Spacy, instead of single entity-tag PERSON.
\begin{table}[t]
\caption{Statistics for Datasets 2 and 3}
\label{tab:datastats}
\begin{tabular}{l c c c c}
    \toprule
    \textbf{Dataset} &\textbf{\#Instances} &\textbf{Avg. \#words in target} & \textbf{\#Target vocabulary}\\
    \midrule
    \textsc{Triple2Text} & 33188424 & 3.45 & 5594\\
    \textsc{MorphKey2Text-V1} & 9481470 & 9.74 & 876153\\
    \textsc{MorphKey2Text-V2} & 9346617 & 8.51 & 477302 \\
\end{tabular}
\end{table}

\subsection{Dataset 3: Knowledge Base Triples to Text (\textsc{Triple2Text})}
For this, a large number of triples and corresponding sentential forms are gathered from the following resources. (i) \textit{Yago Ontology:} 6198617 parallel triples and sentences extracted from Yago \cite{suchanek2007yago}. Our improvised NER, discussed in \emph{Sec.} \ref{sec:canonical} is used for getting tags for entities in the triples. (ii) \textit{OpenIE on WikiData:} 53066988 parallel triples and sentences synthesized from relations from Reverb Clueweb \cite{banko2007open} and all possible combinations of NE Tags. (iii) \textit{VerbNet:} 149760 parallel triples and sentences synthesized from verbs (in the first person singular form) from VerbNet \cite{Schuler:2005:VBC:1104493} and possible combinations of NE Tags. For all the knowledge resources considered for this dataset, concatenation of the elements in the triples yielded simple sentences, hence there was no manual effort needed for creation of this dataset. 

Various statistics for dataset 2 and 3 are presented in \textit{Table.} \ref{tab:datastats}. For training the TextGen systems, the datasets were randomly divided into train, valid and test splits of 80\%:10\%:10\%. 

\section{Experiments}
The simple language generator in \emph{Sec.} \ref{sec:simplegen} require training \texttt{seq2seq} networks using the \textsc{MorphKey2Text} (\textsc{v1} and \textsc{v2}) and the \textsc{Triple2Text} datasets. For this we use the \textsc{OpenNMT} framework in PyTorch, using the default hyperparameter settings. The best epoch model is chosen based on accuracy on the validation split of the above datasets. Once these modules are trained, they are used in inference mode in our pipeline.

Through experiments, we show the efficacy of our proposed system on \textsc{WikiTablePara} and other public \textit{data-to-text} benchmark datasets even though it is not trained on those datasets. Additionally, we also assess the generalizability of our and other existing end-to-end systems  in unseen domains. We use BLEU-4, METEOR, ROUGE-L and \textit{Skip-Thoughts} based Cosine Similarity (denoted as STSim) as the evaluation metrics\footnote{\scriptsize{\url{https://github.com/Maluuba/nlg-eval}}}. We also perform a human  evaluation study, where a held-out portion of the test data is evaluated by linguists who assign scores to the generated descriptions pertaining to \textit{fluency}, \textit{adequacy} and \textit{coherence}. Mainly, we try to answer the following research questions through our empirical study:\\
\begin{enumerate}
    \item \textbf{Can other existing end-to-end systems adapt to unseen domains?} For this, we consider two pre-trained representative models: (a) \textsc{WikiBioModel} \cite{N18-1139} - A neural model trained on the \textsc{WikiBio} dataset \cite{lebret2016neural}, and (b) \textsc{WebNLGModel}\footnote{\scriptsize{\url{http://webnlg.loria.fr/pages/baseline.html}}} - A \texttt{seq2seq} baseline trained on the \textsc{WebNLG} dataset \cite{dbpedia16,nlg-micro17}. These models are tested on the \textsc{WikiTablePara} dataset which is not restricted to any particular domain. Additionally, they are also tested on two popular tuple-to-text datasets such as \textsc{E2E} \cite{W17-5525} and \textsc{WikiTableText} \cite{bao2018table}. Thus, the performance of the existing systems can be assessed on wide variety of domains which may not have been present in the datasets used for developing the systems.
    \item \textbf{How well our system adapts to new domains?} We evaluate our proposed system also on the table-to-descriptions \textsc{WikiTablePara} benchmark dataset to contrast the performance with the above pretrained models. Additionally, we also assess our system on \textit{related (table-to-text summarization)} datasets: (1) \textsc{WebNLG}, (2) \textsc{WikiBio}, (3) \textsc{WikiTableText}, and (4) \textsc{E2E}. The \textsc{WikiTableText} dataset, like ours, is also derived from WikiTables. However, it contains only tabular-rows and their summary in one sentence. The generation objective becomes different from ours, as it does not require  paragraph level operations such as compounding and coreference resolution. Therefore, for brevity, we only report our system's performance on the dataset without further analysis.
    \item \textbf{How interpretable is our approach?} By leveraging the modularity of our system, we would analyze the usefulness of major components in the proposed system and perform error analysis.
\end{enumerate}

\subsection{Experimental Setup} We now discuss how the various systems are configured for evaluation on multiple datasets. 

\begin{itemize}
    \item \textbf{\textsc{Proposed System}} : Our proposed system is already designed to work with the format of the \textsc{WikiTablePara} dataset. Each table in the dataset is converted to $M \times (N-1)$ canonical triples leading to the output table description (refer \textit{Sec.} \ref{sec:canonical}). To test our system for other input types such as \textit{Knowledge Graphs} and \textit{Key-Value} dictionaries, we use the \textsc{WebNLG} and \textsc{WikiBio} datasets respectively. From \textsc{WikiBio} dataset, JSONs containing $N$ Key-Value pairs \emph{$\langle$key1:value1, key2:value2, ... , keyN:valueN$\rangle$} are converted to $N-1$ triples. Each triple is in the form $\langle  value1$, $keyI$, $valueI\rangle$, where $I \neq 1$. It is assumed that the first key is the primary key and typically contains names and other keywords for identifying the original wikipedia infobox.  For \textsc{WebNLG} dataset, the triples in a group are directly used by our system to produce the output. For the  \textsc{WikiTableText} dataset, which contains one tuple per instance, each input is converted into $N-1$, triples, in similar manner as the \textsc{WikiTablePara} dataset. For the \textsc{E2E} dataset, each instance already is in triple-to-text form, and is used as it is.
    
    \item \textbf{\textsc{WebNLGModel}} : The \textsc{WebNLGModel} is designed to be trained and tested on the \textsc{WebNLG} dataset. An already trained \textsc{WebNLGModel} model (similar to the one by \newcite{nlg-micro17}) is evaluated on \textsc{WikiTablePara} and \textsc{WikiBio} datasets. For \textsc{WikiTablePara} dataset, we convert every table to $M \times (N-1)$ triples. For each triple, the model infers a sentence and sentences for all the triples representing a table are concatenated to produce a paragraph description. For the \textsc{WikiBio} dataset, each JSON is converted to $N-1$ triples for $N$ key-value pairs, which are then passed to the model for final output. Tuples in \textsc{WikiTableText} dataset are converted to $N-1$ triples and instances in \textsc{E2E} dataset, which are already in triple-to-text are used directly without any transformation.
    
    \item \textbf{\textsc{WikiBioModel}} : The \textsc{WikiBioModel} is designed to get trained and tested on the \textsc{WikiBio} dataset that contains Key-Value pairs at the input side and summaries at the output. An already trained model (similar to the one by \newcite{N18-1139}) is evaluated on \textsc{WikiTablePara} and \textsc{WebNLG} datasets. For the \textsc{WikiTablePara} dataset, we convert every table to $M \times (N-1)$ jsons in \textsc{WikiBio} format. Each JSON contains a pair of Key-Value pairs, where the first Key-Value pair always represents the primary-key and its corresponding entry in the table (hence, $N-1$ JSONs are produced). The inferred sentences for all $M \times (N-1)$ jsons from the model are concatenated to produce the required paragraph description. For the \textsc{WebNLG} dataset, each triple is converted to a JSON of a pair of Key-Value pairs. A triple $\langle e_1~,~r~,~e_2\rangle$ is converted to a JSON format of $\{default\_key:e_1~,r:e_2\}$ (the defalt key is set to \textit{``name''}). For each instance in the \textsc{WebNLG} dataset, sentences are inferred for all the triples belonging to the instance, and they are concatenated to produce the final output. Inputs from \textsc{WikiTableText} and \textsc{E2E} datasets are converted to JSON as explained above.
\end{itemize}

Please note that both \textsc{WikiBioModel} and \textsc{WebNLGModel} are capable of processing single and multi-tuple inputs. For our dataset, we try giving these models inputs in both single and multi-tuple format. In single-tuple input mode, the model processes one triple at a time and produces a sentence; the sentences are concatenated to produce paragraphs. In multi-tuple mode, all triples extracted from a single row of the table are simultaneously passed to the model as input. The model variants with subscript \textit{``M''} represent these cases in the result tables. For the above evaluations, only the test splits for \textsc{WikiBio} and \textsc{WebNLG} datasets are used, whereas there is no train:test split for the \textsc{WikiTablePara} dataset (the entire  dataset is used for evaluation). The results for these are summarized in \emph{Tables} \ref{tab:allmodelourdatatable} and \ref{tab:crossdataevaluation}.

\subsubsection{Ablation Study} Apart from comparing our system with the  existing ones, we also try to understand how different stages of our pipeline contribute to the overall performance. For such an ablation study, we prepare the different variants of the system based on the following two scenarios and compare their performance against that of the complete system. 

\begin{itemize}
    \item Instead of using the ensemble (\textsc{Ranker}), each participating TextGen systems \textit{viz.} \textsc{triple2text}, \textsc{morphkey2text v1} and \textsc{morphkey2text v2} are treated as separate system. The intention is to show the advantage of using an ensemble of generators and the ranking mechanism. 
    \item Language enrichment modules such as compounding and coreference replacement modules are removed both individually and together. Simple sentences are just concatenated to produce the table descriptions. The intention is to test our hypothesis that \textit{removing such modules will make the generated paragraphs somewhat incoherent and deviate from constructs produced by humans, thereby resulting in a reduced system performance}. 
\end{itemize}
\section{Results and Discussion}
\label{sec:results}
\emph{Table} \ref{tab:allmodelourdatatable} illustrates how the various pretrained models fare on the \textsc{WikiTablePara} benchmark dataset compared to our proposed system. We observe that the end-to-end \textsc{WebNLGModel} does better that \textsc{WikiBioModel}. However, our proposed system clearly gives the best performance, demonstrating the capability of generalizing in unseen domains and structured data in a more complex form such as multi-row and multi-column table.

It may be argued that although the proposed model is not trained on parallel data, it takes advantage from the fact that the textual resources used for development come from the same sources as the test data (\textit{i.e.,} Wikipedia). Thus, the better performance can be  attributed to having a better vocabulary coverage (covering more entities, verbs, nouns \textit{etc.}) which \textsc{WebNLGModel} and \textsc{WikiBioModel} are deprived of. This is, however, not true because of two reasons: (1) the \textsc{WikiBio} and \textsc{WebNLG} datasets use information from Wikipedia (in the form of Infoboxes and DBPedia entries), or (2) use pre-trained Glove embeddings \cite{pennington2014glove}, which offers a much richer vocabulary than what is considered in our setting. Hence, it is evident that the performance of these baseline systems is low on \textsc{WikiTablePara} dataset not because of vocabulary \textit{unseenness} but for the very fact that these systems are rigid with respect to the language patterns seen in the data they are trained on.

It may also seem unfair to compare standalone systems like \textsc{WikiBioModel} and \textsc{WebNLGModel} with an ensemble model like ours, as the latter may have infused more knowledge because of the inclusion of supporting modules. Again, this is not entirely true. The \textsc{WikiBioModel} under consideration is more sophisticated than a vanilla sequence-to-sequence model and employs attention mechanisms at various levels to handle intricacies in content selection and language generation \cite{N18-1139}. The \textsc{WebNLGModel} employs various normalization and post-processing steps to adapt to newer domains and language patterns. In sum, these models are capable of handling nuances in data-to-text generation and, hence, deem fit for comparison.

\begin{table}[t]
\caption{Various models on \textsc{WikiTablePara} dataset}
\label{tab:allmodelourdatatable}
\begin{tabular}{l c c c c}
    \toprule
    \textbf{System}&\textbf{BLEU}&\textbf{METEOR}&\textbf{ROUGE-L}&\textbf{STSim}\\
    \midrule
    \textsc{WikiBioModel}&0.0&15.5&14.8&64.1\\
    \textsc{WebNLGModel}&7.9&24.8&27.9&78.2\\
    \textsc{Proposed}&\textbf{33.3}&\textbf{39.7}&\textbf{64.1}&\textbf{86.5}\\
\end{tabular}
\end{table}

\begin{table}
\caption{Evaluation of all models across datasets (domains). Suffix \emph{M} represents multi-tuple input.}
\label{tab:crossdataevaluation}
\begin{tabular}{l l c c c c c}
    \toprule
    \textbf{Model}&\textbf{Dataset}&\textbf{BLEU}&\textbf{METEOR}&\textbf{ROUGE-L}&\textbf{STSim}\\
    \midrule
    \multirow{5}{*}{Proposed}&\textsc{WebNLG}&24.8&34.9&52.0&82.6\\
    &\textsc{WikiTableText}&12.9&33.6&37.1&73.2\\
    &\textsc{WikiBio}&2.5&17.6&19.3&72.9\\
    &\textsc{E2E}&6.6&27.1&29.2&71.1\\
    &\textsc{WikiTablePara} &33.3&39.7&64.1&86.5\\
    \midrule
    \multirow{4}{*}{\textsc{WikiBioModel}}&\textsc{WebNLG}&2.8& 16.9& 26.4&72.1 \\
    &\textsc{WikiTableText}&1.3 &10.5 &21.5 &66.5 \\
    &\textsc{E2E} &1.3& 9.0& 22.7& 61.6\\
    &\textsc{WikiTablePara} &0.0&15.5&14.8&64.1\\
    &\textsc{WikiTablePara}$_{M}$ & 0.0 & 10.3 & 13.7 & 65.8 \\
    \midrule
    \multirow{4}{*}{\textsc{WebNLGModel}}&\textsc{WikiTableText} &3.6 &16.5 &25.2 &68.9 \\
    &\textsc{WikiBio} &1.6 &9.3 &18.6 & 69.4\\
    &\textsc{E2E}&2.1 &13.2 &19.0 & 66.0\\
    &\textsc{WikiTablePara} &7.9&24.8&27.9&78.2\\
    &\textsc{WikiTablePara}$_{M}$ &0.5 & 20.0 & 26.1 & 75.6 \\
\end{tabular}
\vspace{-0.5cm}
\end{table}
\begin{table}[t]
\caption{Ablation study: Performance of individual TextGen systems with the ensemble system enabled with \textsc{Ranker}. Here, MKT denotes \textsc{MorphKey2Text}, CP refers to the Compounding Module and CR means Coreference Replacement. The symbols `$+$' and `$-$' signify ``with'' and ``without'' respectively.}
\label{tab:fullablation}
\begin{tabular}{@{}llllll@{}}
\toprule
                         &  & \textsc{Ranker} & \textsc{MKTv1} & \textsc{MKTv2} &
                         \textsc{triple2text} \\
                         \midrule
\multirow{4}{*}{BLEU}    & -CP-CR & 17.7&16.2&14.9&20.3  \\
                         & +CP-CR & 30.1&29.7&27.9&30.3 \\
                         & -CP+CR & 29.6&29.3&28&27.8  \\
                         & +CP+CR & \textbf{33.3}&\textbf{30.6}&\textbf{29.4}&\textbf{30.5}  \\
                         \midrule
\multirow{4}{*}{METEOR}  & -CP-CR & 33.1&33.4&32.6&31.6   \\
                         & +CP-CR & 38.8&39.3&38.4&36.7  \\
                         & -CP+CR & 37.1&37&37&34.8  \\
                         & +CP+CR & \textbf{39.7}&\textbf{38.1}&\textbf{38.1}&\textbf{35.4} \\
                         \midrule
\multirow{4}{*}{ROUGE-L} & -CP-CR & 50.2&51&49.1&50.6  \\
                         & +CP-CR & 61.8&62.3&60.7&61 \\
                         & -CP+CR & 59.2&59&58.7&58.4  \\
                         & +CP+CR & \textbf{64.1}&\textbf{63.9}&\textbf{62.2}&\textbf{62.2} \\
                         \midrule
\multirow{4}{*}{STSim}   & -CP-CR & 44.1&40.2&40.2&57.8 \\
                         & +CP-CR & 85.3&85.6&85.2&83.3  \\
                         & -CP+CR & 82.3&82&82&79.8  \\
                         & +CP+CR & \textbf{86.5}&\textbf{85.9}&\textbf{85.9}&\textbf{83.9} \\ 
\end{tabular}
\vspace{-0.3cm}
\end{table}

\emph{Table} \ref{tab:crossdataevaluation} shows the performance of our proposed system on the test splits of various datasets (including the whole \textsc{WikiTablePara} dataset). The performance measures (especially the STSim metric) indicate that our system can be used as it is for other input types coming from diverse domains. Despite the fact that the \textsc{WikiTableText}, \textsc{WebNLG} and \textsc{WikiBio} datasets are summarization datasets and are not designed for complete description generation, our system still performs reasonably well, without having been trained on any of these datasets. It is clearly observed that the existing end-to-end models such as \textsc{WebNLGModel} and \textsc{WikiBioModel} exhibit inferior cross-domain performance compared to our system\footnote{Please note that \textsc{WikiBioModel} trained on \textsc{WikiBio} dataset (in-domain) would have considerably higher evaluation scores (refer \newcite{N18-1139}); the same holds for the \textsc{WebNLGModel} \cite{nlg-micro17}. Since our objective is to highlight cross-domain performance (where testing is done on datasets different from training data), the in-domain results are not discussed for brevity.}. For example, our system attains BLEU scores of 24.8 and 2.5 on \textsc{WebNLG} and \textsc{WikiBio} datasets respectively, whereas, the \textsc{WikiBioModel} performs with a BLEU score of 2.8 (with a reduction of 89\%) on the \textsc{WebNLG} dataset and the \textsc{WebNLGModel} performs with a BLEU score of 1.6 (with a reduction of 36\%) on \textsc{WikiBIO} dataset. For other datasets such as \textsc{WikiTableText} and \textsc{E2E}, on which none of the proposed or comparison systems are trained on, our system's performance is significantly better than the comparison systems. For the \textsc{E2E} dataset, we observe that our system's outputs convey similar semantics as the reference texts but have considerable syntactic differences. For examples, the triples $\langle\textbf{Taste of Cambridge} \ \emph{eat type} \ \textbf{restaurant}\rangle$, and $\langle\textbf{Taste of Cambridge} \ \emph{customer rating} \ \textbf{3 out of 5}\rangle$ are translated to \emph{``Taste of Cambridge is an eat type of restaurant and has a customer rating of 3 out of 5.''} by one of our model variant, but the reference text is \emph{``Taste of Cambridge is a restaurant with a customer rating of 3 out of 5.''} This may have affected the BLEU scores; the METEOR and semantic relatedness scores are still better.

We performed ablation on our proposed system at multiple levels; \emph{Table} \ref{tab:fullablation} shows the performance of individual simple language generation systems and also the performance of the ranker module. The results suggest that ranker indeed improves the performance of the system. To measure the effectiveness of our proposed sentence compounding and coreference replacement modules, we replaced these modules with a simple sentence concatenation module. As observed in the same table, the performance of the system degrades compared to when compounding and coreference replacement modules are individually used. Best results are obtained when both the modules are activated. One of the possible reasons is that a simple sentence concatenation results in generated paragraphs having more redundant occurrences of entity terms and phrases, which all of the evaluation metrics tend to penalize heavily. Overall, this study goes to show that the enrichment modules indeed play an important role, especially when it comes to paragraph description generation.
\begin{figure}[t]
 \centering
 \includegraphics[width=1.0\textwidth, height=16.5cm, keepaspectratio=true]{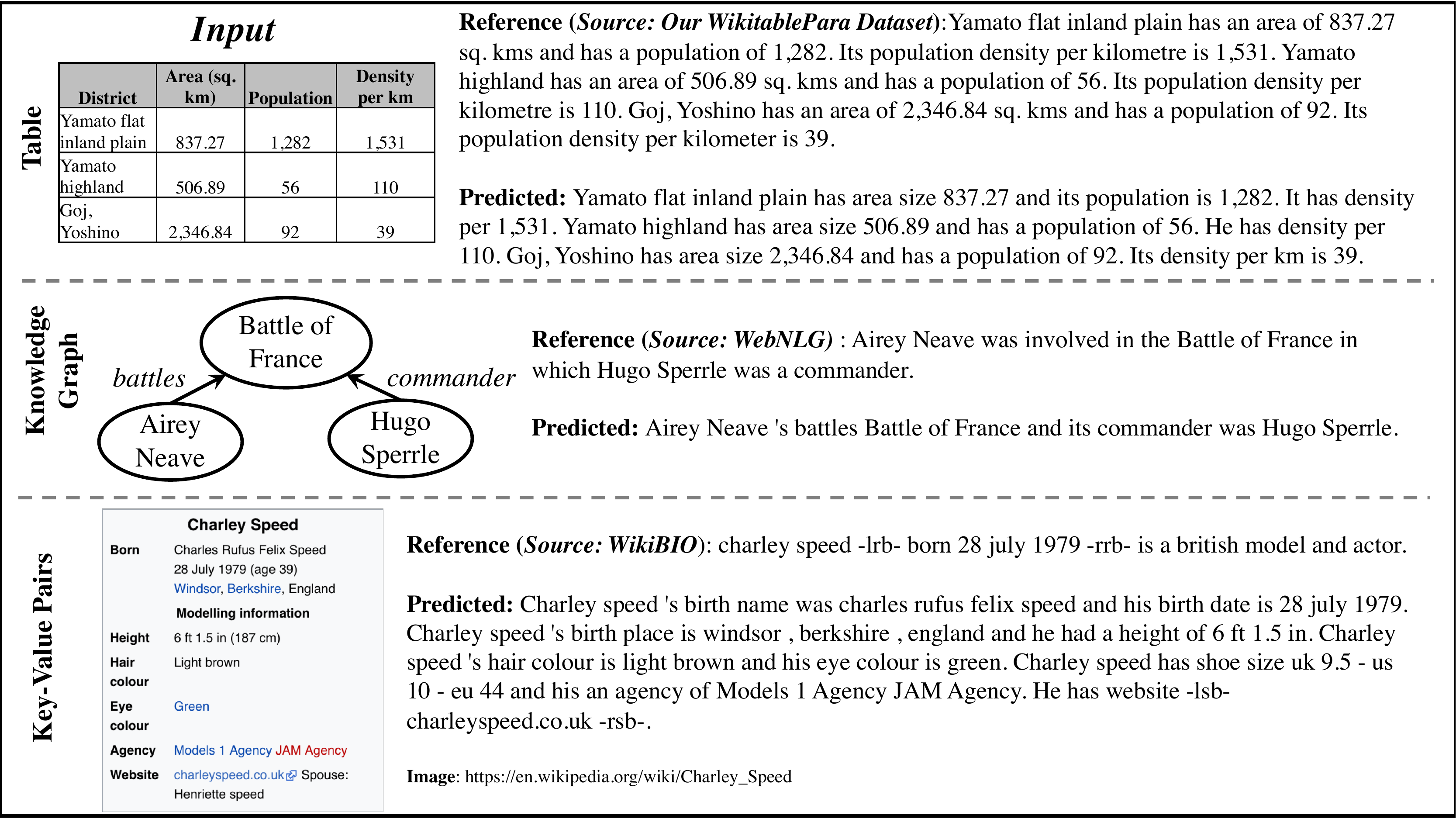}
 \caption{Examples of generated descriptions by our proposed system on different datasets.}
 \label{fig:ouranecdote}
 \end{figure}
\begin{table}[t]
 \caption{Human evaluation using 50 samples from the \textsc{WikiTablePara} dataset. The fluency, adequacy and coherence scores are averaged across evaluators and instances. Evaluator correlation is the Pearson Correlation which shows the agreement between evaluators.}
 \label{tab:humanevaluation}
 \begin{tabular}{l c c c}
     \toprule
     \textbf{System}&\textbf{Fluency}&\textbf{Adequacy}&\textbf{Coherence}\\
     \midrule
     \textsc{WikiBioModel}& 1.44 & 1.24 & 1.08 \\
     \textsc{WebNLGModel}& 2.04 & 2.05 & 1.66\\
     \textsc{Proposed}& \textbf{3.29}&\textbf{4.20}&\textbf{3.72} \\
     \midrule
     \textit{\textsc{GOLD-standard}} & \textit{4.53}&\textit{4.78}&\textit{4.59} \\
     \midrule
     \textit{Evaluator Correlation} & \textbf{\textit{0.74}} & \textbf{\textit{0.80}} & \textbf{\textit{0.76}} \\
 \end{tabular}
 \vspace{-0.5cm}
\end{table}

\subsection{Human Evaluation}
Since quantitative evaluation metrics such as BLEU and Skip-thought similarity are known to have limited capabilities in judging sentences that are correct but different from the gold-standard reference, we perform a human evaluation study. For this, the first $50$ instances from the \textsc{WikiTablePara} dataset were selected. For each instance, the table, the reference paragraph, and outputs from our proposed system, \textsc{WikiBio} and \textsc{WebNLG} models were shuffled and shown to four linguists. They were instructed to assign three scores related to fluency, adequacy and coherence of the generated and gold-standard paragraphs. The minimum and maximum scores for each category are 1 and 5 respectively. \textit{Table.} \ref{tab:humanevaluation} reports the evaluation results. While it was expected that the gold-standard output would get maximum average scores in all aspects, the scores for our proposed systems are quite superior to the existing systems and are also sometimes close to those for the gold standard paragraphs. This shows that a modular approach like ours can be effective for generating tabular descriptions. Moreover, the average Pearson Correlation coefficient values  for each scores across systems and evaluator-pairs are high, showing a strong inter-evaluator agreement. 

On manual inspection of the descriptions generated by our system across datasets (some examples are shown in \emph{Fig.} \ref{fig:ouranecdote}), we find that our system gives a promising performance in addition to the quantitative evaluation metrics mentioned before.
\subsection{Effectiveness of the Individual Modules}
\begin{figure}[ht]
\centering
\includegraphics[width=0.5\linewidth, keepaspectratio=true]{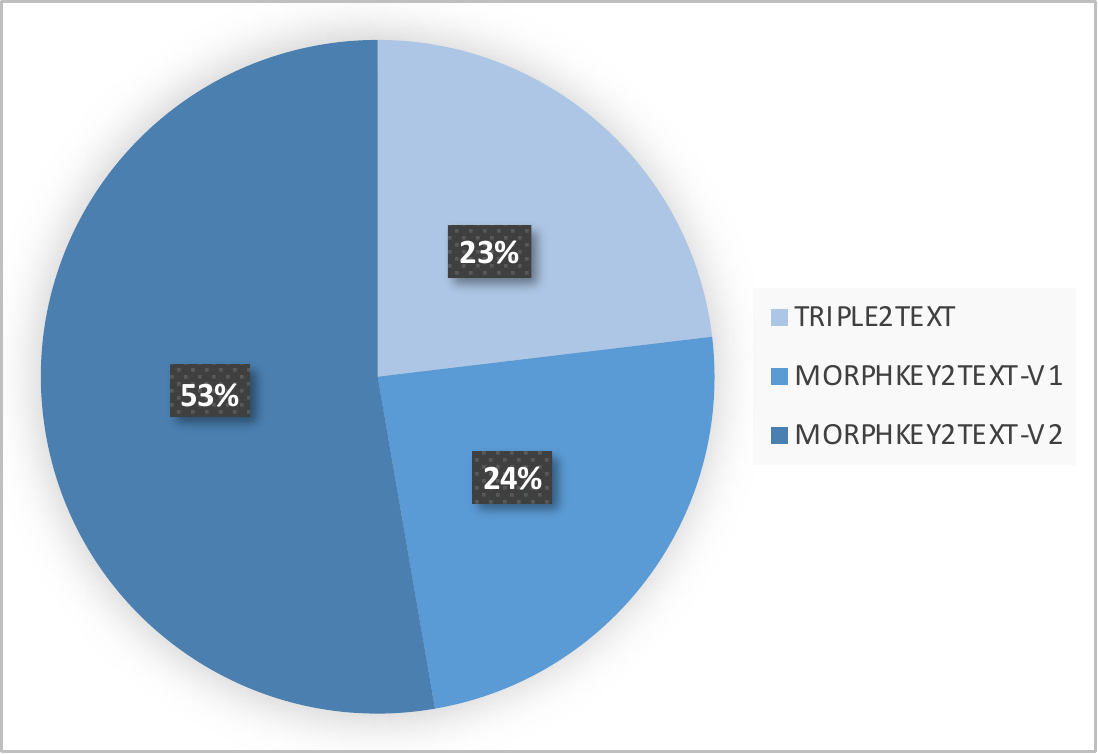}
\caption{Contribution of each TextGen system in term of percentage of times the output of these systems were selected by the ranker for the \textsc{WikiTablePara} dataset.}
\label{fig:pie}
\end{figure}

We also examine if, for TextGen, using an ensemble of generators followed by a ranking mechanism was effective. We intend to study if all the participating systems were chosen by the ranker for a significant number of examples. \textit{Fig.} \ref{fig:pie} shows the percentage of the times the output of the three TextGen systems were selected by the ranker. As we can see, all systems are significantly involved in producing the correct output in the test data. However, the  \textsc{triple2text} system is selected fewer number of times than the other two systems. This is a positive result as the \textsc{triple2text} system requires data obtained from specific resources such as OpenIE, and Yago as opposed to the \textsc{morphkey2text} systems that require just a monolingual corpus. 

\vspace{-0.5cm}
\subsection{Error Analysis}
Since our system is modular, we could inspect the intermediate outputs of different stages and perform error-analysis. We categorize the errors into the following: 
\begin{itemize}
    \item \textbf{Error in Tagging of Entities:} One of the crucial steps in the canonicalization stage is tagging the table entries. Our modified NE taggers sometimes fail to tag entities primarily because of lack of context.  For example, the original triple in our dataset $\langle$ \textit{Chinese Taipei, gold medals won, 1} $\rangle$ is converted to a triple $\langle$ \textit{\textbf{UNK}, gold medals won, \textbf{CARDINAL}} $\rangle$. Because of the wrong NE-tagging of the entity Chinese Taipei, the text generation stage in the pipeline did not get enough context and failed to produce a fluent output as shown below, 
    \vspace{-0.2cm}
    \begin{quote}
        \textit{Chinese Taipei's gold medals have been won by 0.}
        \vspace{-1ex}
    \end{quote}
    This error affected all the subsequent stages. While it is hard to resolve this with existing NLP techniques, maintaining and incrementally building gazetteers of domain specific entities for look-up based tagging can be a temporary solution. 
    \item {\textbf{Error in the TextGen:}} We observe that all the TextGen systems, discussed in \textit{Sec.} \ref{sec:triple2text} are prone to syntactic errors, which are mostly of types \textit{subject-verb disagreement}, \textit{noun-number disagreement}, article and preposition errors. An example of such an erroneous output is shown below:
    \begin{quote}
    \textit{Republican 's active voters is 13,916. Republican was inactive in voters 5,342}
    \vspace{-1ex}
    \end{quote}
    We believe such errors can be avoided by adding more training examples, judiciously prepared from large scale monolingual data from different domains.
    \item {\textbf{Error in Ranking:}} This error impacts the performance of our system the most. We consistently observe that even though one of the individual systems is able to produce fluent and adequate output, it is not selected by the ranker module. In the hindsight, scorers based on simple language models and content-overlaps (\textit{Eq.} \ref{eq:score}) are not able to capture diverse syntactic and semantic representations of the same context (\textit{e.g.,} passive forms, reordering of words). Moreover, language models are known to capture N-gram collocations better than the overall context of the sentences, and tend to penalize grammatically correct sentences more than the incorrect sentences that have more \textit{likely} collocations of N-grams. Furthermore, longer sentences are penalized more by the language model than shorter ones. To put this into perspective, consider the following example from our dataset. For the input triple, $\langle$\textit{Bischofsheim, building type, Station building}$\rangle$, the output from the TextGen systems are as follows:
    \begin{quote}
        \textbf{\textsc{triple2text:}} 
        \textit{Bischofsheim has building type Station Building.} 
        \textbf{\textsc{morphkey2text-v1:}} \textit{Bischofsheim's building is a type of Station Building.}
        \textbf{\textsc{morphkey2text-v2:}} \textit{Bischofsheim is a building type of Station Building.}
        \vspace{-1ex}
    \end{quote}
    
    The ranker unfortunately selects an imperfect output produced by the \textsc{morphkey2text-v2} system. We believe that the presence of highly probable bigrams such as \emph{building type} and \emph{type of} would have bolstered the language model score and, eventually the overall score. A possible solution to overcome this would be to train neural knowledge language models \cite{nklm16} that not only considers contextual history but also factual correctness of the generated text. Gathering more monolingual data for training such models may help as well. 
    
    \item {\textbf{Error in Coreference Determination:}} Error in coreference determination happens due to two reasons : (a) The entities are incorrectly tagged (\textit{e.g.,} a \textit{PERSON} is mis-tagged as \textit{ORG}, leading to a wrong pronominal anaphora.), and (b) The gender of the entity is incorrectly classified (\textit{e.g.,} Esther Ndiema's nationality is Kenya and \underline{\textit{his}} rank is 5). While improving the tagger is important for this and the overall system, the gender detector could be improved through more training data and better tuning of hyperparameters. The current module does have limitations due to the fact that it is based on a very small number of heuristics and relies on data-driven POS-taggers and gender predictors, which may not provide accurate information about the number and gender of the mentions. For example, for an entity ``Mariya Papulov'', even though POS tagger and the canonicalized entity tag (PERSON) help determine the number of the coreference correctly, the gender predictor assigns the gender tag as male. This results in a wrong co-reference assignment. 

\end{itemize}    
A deeper issue with the sentence enrichment modules is that they are agnostic of the sentence order. If the TextGen systems do not provide sentences in an  appropriate order to these modules, the cohesiveness of the generated paragraph is compromised. For example, the output from our system \emph{``Melania Corradini played for Italy and was on the run of distance 54.72 KMs. She had the rank of 5.''} provides a  less natural feel than \emph{``Melania Corradini played for Italy and had the rank of 5. She was on the run of distance 54.72 KMs''}. This clearly calls for a technique to determine the optimal order of sentences to ensure more naturalness in the output.
    
We would also like to point out that language enrichment through simple concatenation and heuristic based replacement is a  rudimentary solution. Better solutions for compounding and producing coherent paragraphs may involve syntactic analysis and restructuring of sentences \cite{D17-1064} and discourse aware coherent generation \cite{D17-1064,kibble2004optimizing,N18-1016}.

\section{Related Work}
\label{sec:related}
Data-to-text NLG has received a lot of attention in recent times, especially due to the increasing demands of such systems for industrial applications. Several such systems are based on rule-based, modular statistical and hybrid approaches and are summarized by \newcite{N18-1139}. Recently, end-to-end neural generation systems have been preferred over others. Some of the most recent ones are based on the \textsc{WikiBio} dataset \cite{lebret2016neural}, a dataset tailor-made for summarization of structured data in the form of key-value pairs. Such systems include the ones by \newcite{lebret2016neural}, who use conditional language model with copy mechanism for generation, \cite{liu2017table} who propose a dual attention Seq2Seq model,  \newcite{N18-1139} who employ gated orthogonalization along with dual attention, and \newcite{bao2018table} who introduce a flexible copying mechanism that selectively replicates contents from the table in the output sequence. Other systems revolve around popular datasets such as \textsc{WeatherGov} dataset \cite{liang2009learning,N18-2098}, \textsc{RoboCup} dataset \cite{chenmooney08}, \textsc{RotoWire} and \textsc{SBNation} \cite{D17-1239}, and \textsc{WebNLG} dataset \cite{nlg-micro17}. Recently \newcite{bao2018table} and \cite{W17-5525} have introduced a new dataset for table/tuple to text generation and both supervised and unsupervised systems \cite{fevry2018unsupervised} have been proposed and evaluated against these datasets. The objective of creating such datasets and systems is, however,  entirely different from ours. For instance, \newcite{bao2018table}'s objective is to generate natural language summary for a \textbf{region} of the table such as a row, whereas, we intend to translate the complete table into paragraph descriptions. This requires additional discourse level operations (such as sentence compounding and coreference insertion). The dataset also contains tabular rows at the input side and summaries at the output side. Since the objective is summarization of a tabular region, a fraction of the entries are dropped and not explained, unlike ours that aims to translate the complete table into natural language form.
 
It is worth noting that recent works for keyword to question generation \cite{reddy2017}, and set-to-sequence generation \cite{44871} can also act as building blocks for generation from structured data. However, none of these works consider the morphological  and linguistic variations of output words as we consider for simple language generation. Also, the work on neural knowledge language model \cite{nklm16} incorporates facts towards a better language model, which is a different goal from ours as we attempt to describe the full table of facts in natural language in a coherent manner. However, irrespective of the generation paradigms we use, the bottom line remains the same - modular data driven approaches like ours can produce robust and scalable solutions for data-to-text NLG.

It is also worth noting that there exist well-studied information extraction (IE) techniques to obtain tuples from sentences which could be employed to prepare additional parallel training data for improving any data-to-text NLG solution. Here, the tuples can be used as source and the original (or preprocessed) sentences can be used as target for training supervised generators. Popular IE techniques include Open Information Extraction \cite{banko2007open} and Open Language Learning for Information Extraction  \cite{schmitz2012open}. These systems leverage POS taggers and dependency parsers to extract relation tuples and are in line with our method for data generation. However, they do not consider lexical and morphological aspects of the sentences considered as we did for \textsc{MorphKey2Text}. From the domain of relation extraction, works such as \cite{mintz2009distant}, that extract training datasets for relation annotation using distant supervision techniques, can also be improvised and used in our setting.
\vspace{-0.3cm}
\section{Conclusion and Future Directions}
\label{sec:conslusion}
We presented a modular framework for generating paragraph level natural language description from structured tabular data. We highlighted the challenges involved and contended why a modular data-driven architecture like ours could tackle them better as opposed to end-to-end neural systems. Our framework employs modules for obtaining standard representations of tables, generating simple sentences from them and finally combining the sentences to form coherent and fluent paragraphs. Since no benchmark dataset for evaluating discourse level description generation was available, we created one to evaluate our system. Our experiments on our dataset and various other \emph{data-to-text} type datasets reveal that: (a) our system outperforms the existing ones in producing discourse level descriptions without undergoing end-to-end training on such data, and (b) the system can realize good quality sentences for various other input data-types such as \textit{knowledge graphs} in the form of tuples and key-value pairs. Furthermore, the modularity of the system allows us to interpret the system's output better. In the future, we would like to incorporate additional modules into the system for tabular summarization. Extending the framework for multilingual tabular description generation is also on our agenda.

\starttwocolumn
\bibliography{cl2019}
\bibliographystyle{compling}
\end{document}